\documentclass[conference]{IEEEtran}
\usepackage{hyperref}
\usepackage{color}
\usepackage{float}
\usepackage{tcolorbox}
\usepackage{subcaption}
\usepackage{authblk}
\usepackage{bm}
\usepackage{amsmath}
\usepackage{lettrine}

\usepackage{float}
\usepackage{stfloats}
\usepackage[ruled,vlined]{algorithm2e}
\usepackage{times}
\usepackage[para]{threeparttable}
\usepackage{array,booktabs,longtable,tabularx}
\usepackage{etoolbox}
\usepackage{authblk}
\usepackage[noadjust]{cite}
\usepackage{tabu}
\usepackage{tikz}
\newcommand*\circled[1]{\tikz[baseline=(char.base)]{
            \node[shape=circle,draw,inner sep=0.8pt] (char) {#1};}}
\usepackage{multirow}
\usepackage{orcidlink}

\usepackage{acronym}
\acrodef{CMOS}[CMOS]{Complementary Metal–Oxide–Semiconductor}
\acrodef{BMI}[BMI]{Brain-Machine Interface}
\acrodef{ANN}[ANN]{Artificial Neural Network}
\acrodef{IMC}[IMC]{In-Memory Computing}
\acrodef{RRAM}[RRAM]{Resistive Random-Access Memory}
\acrodef{EEG}[EEG]{Electroencephalogram}
\acrodef{ML}[ML]{Machine Learning}
\acrodef{FP}[FP]{False Positive}
\acrodef{FN}[FN]{False Negative}
\acrodef{TN}[TN]{True Negative}
\acrodef{TP}[TP]{True Positive}
\acrodef{ROC}[ROC]{Receiver Operating Characteristic}
\acrodef{ICV}[ICV]{Intracluser Variance}
\acrodef{SNR}[SNR]{Signal-To-Noise Ratio}
\acrodef{ADC}[ADC]{Analog-To-Digital Converter}
\acrodef{DAC}[DAC]{Digital-To-Analog Converter}
\acrodef{SA-ADC}[SA-ADC]{Successive-Approximation ADC}
\acrodef{S/H}[S/H]{Sample-and-Hold}
\acrodef{SAR}[SAR]{Successive Approximation Register}
\acrodef{MSB}[MSB]{Most-Significant-Bit}
\acrodef{DSP}[DSP]{Digital Signal Processor}
\acrodef{FLPA}[FLPA]{Functional-Level Power Analysis}
\acrodef{UWB}[UWB]{Ultra-Wide Band}
\acrodef{POC}[POC]{Point-Of-Care}
\acrodef{PCA}[PCA]{Principal Component Analysis}
\acrodef{FPGA}[FPGA]{Field-Programmable Gate Array}
\acrodef{ZCF}[ZCF]{Zero-Crossing Feature}
\acrodef{DL}[DL]{Deep Learning}
\acrodef{VMM}[VMM]{Vector-Matrix Multiplication}
\acrodef{DWT}[DWT]{Discrete Wavelet Transform}
\acrodef{CAS}[CAS]{Circuits and Systems}
\acrodef{PCC}[PCC]{Pearson Correlation Coefficient}
\acrodef{BEOL}[BEOL]{Back-End-Of-The-Line}
\acrodef{SSOER}[SSOER]{Synthetic Simulations Of Extracellular Recordings}

\title{Toward A Formalized Approach for Spike Sorting Algorithms and Hardware Evaluation}

\author[1]{Tim Zhang\orcidlink{0000-0002-2740-8693}\thanks{\hspace{-1em}\rule{3cm}{0.5pt} \newline \textcopyright  \hspace{1pt} 2022 IEEE. Personal use of this material is permitted. Permission from IEEE must be obtained for all other uses, in any current or future media, including reprinting/republishing this material for advertising or promotional purposes, creating new collective works, for resale or redistribution to servers or lists, or reuse of any copyrighted component of this work in other works.}} 
\author[2]{Corey Lammie\orcidlink{0000-0001-5564-1356}}
\author[2]{Mostafa Rahimi Azghadi\orcidlink{0000-0001-7975-3985}}
\author[3]{Amirali  Amirsoleimani\orcidlink{0000-0001-5760-6861}}
\author[4]{Majid Ahmadi\orcidlink{0000-0001-5781-6754}}
\author[5]{\authorcr Roman Genov\orcidlink{0000-0001-7506-1746}}
\affil[1]{Department of Bioengineering, McGill University, Montreal H3A 0G4, Canada}
\affil[2]{College of Science and Engineering, James Cook University, Queensland 4814, Australia}
\affil[3]{Department of Electrical Engineering and Compute Science, York University, Toronto ON M3J 1P3, Canada}
\affil[4]{Department of Electrical and Computer Engineering, University of Windsor, Windsor, Canada}
\affil[5]{Department of Electrical and Computer Engineering, University of Toronto, Toronto, Canada}
\affil[$\relax$]{Email:$^1$tianyi.zhang4@mail.mcgill.ca,  $^2$\{corey.lammie, mostafa.rahimiazghadi\}@jcu.edu.au, $^3$amirsol@yorku.ca,}
\affil[$\relax$]{
$^4$ahmadi@uwindsor.ca, $^5$roman@eecg.utoronto.ca \vspace{-5mm}}

\IEEEoverridecommandlockouts
\begin{document}
\maketitle

\begin{abstract}
Spike sorting algorithms are used to separate extracellular recordings of neuronal populations into single-unit spike activities. The development of customized hardware implementing spike sorting algorithms is burgeoning. However, there is a lack of a systematic approach and a set of standardized evaluation criteria to facilitate direct comparison of both software and hardware implementations.
In this paper, we formalize a set of standardized criteria and a publicly available synthetic dataset entitled \textit{\ac{SSOER}}, which was constructed by aggregating existing synthetic datasets with varying \acp{SNR}.
Furthermore, we present a benchmark for future comparison, and use our criteria to evaluate a simulated \ac{RRAM} \ac{IMC} system using the \ac{DWT} for feature extraction. Our system consumes approximately (per channel) 10.72mW and occupies an area of 0.66mm$^2$ in a 22nm FDSOI \ac{CMOS} process.
\end{abstract}

\begin{IEEEkeywords}
Spike Sorting, RRAM, IMC, CMOS, DL
\end{IEEEkeywords}
\IEEEpeerreviewmaketitle

\section{Introduction}
\lettrine{E}{lectrophysiology}, extracellular recordings of neuronal populations, has become a cornerstone for neuroscience research due to its ability to measure action potential and neuronal activities in the vicinity of electrodes. However, extracellular recordings are the summation of action potentials fired by a variety of neurons within the recording vicinity, and consequently require decoding. Advances in \ac{CMOS} and emerging \ac{IMC} technologies~\cite{RahimiAzghadi2020}, such as \ac{RRAM}, allow for an exponential increase in number of neurons that can be simultaneously recorded, which calls for development of fast, efficient, and automated spike sorting algorithms to decode the recorded information.

\begin{figure}[!t]
    \centering
    \includegraphics[width=0.9\linewidth]{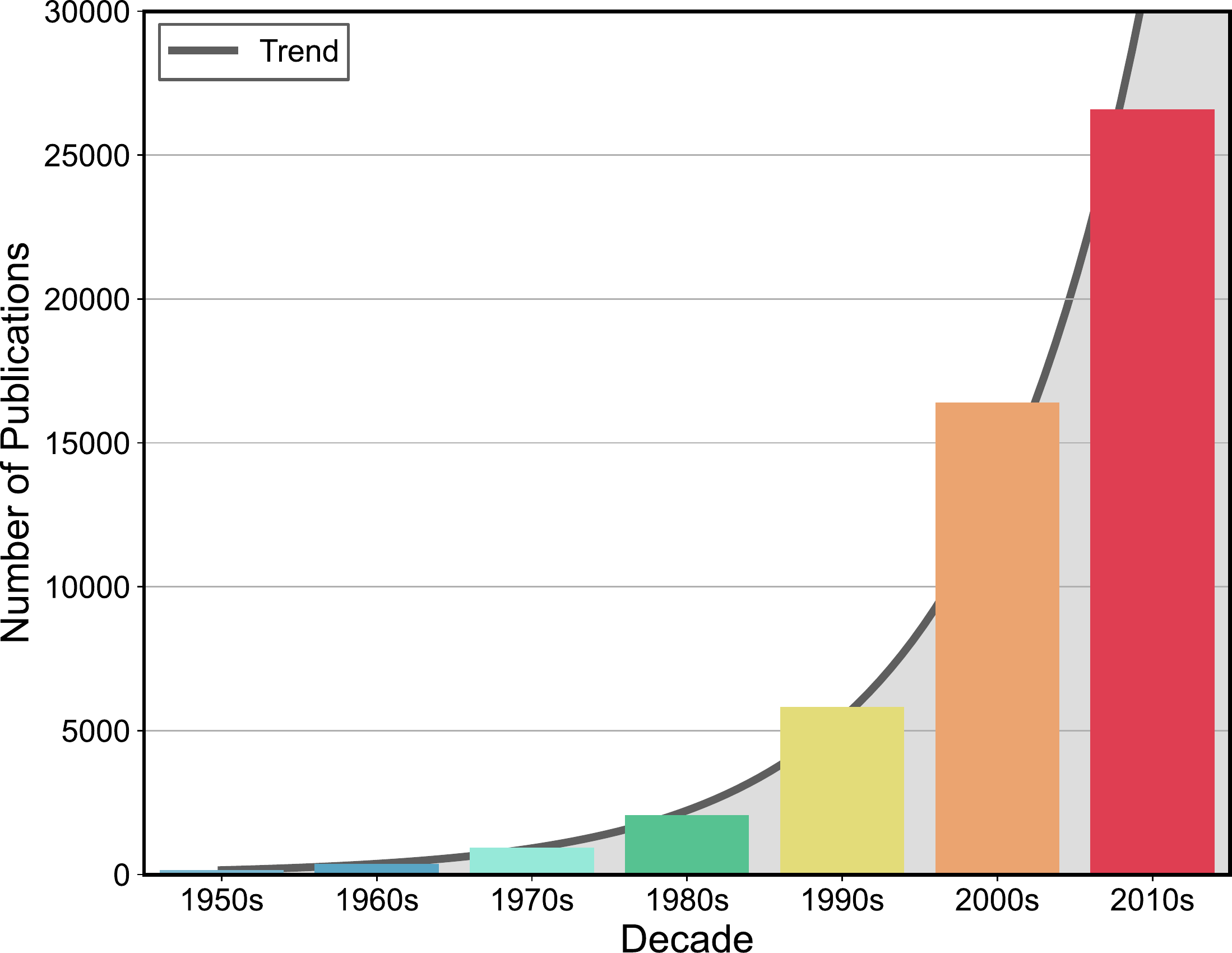}
    \caption{Number of publications related to spike sorting over the decades since its pioneering in the 1950s.}
    \label{fig:number_of_publications}
\end{figure}

Extracellular recordings paired with spike sorting are pre-requisites for both commercial applications, such as \acp{BMI}, which can restore motor functions or damaged sensory functions~\cite{hochberg2006neuronal}, and for research applications, where interactions between different neurons and their network effects that give rise to complex higher order functions such as movement, perception, and memory are studied. Since the pioneering work of the spike sorting field in the 1964~\cite{gerstein1964simultaneous}, the field has attracted an exponentially growing attention from the neuroscience and engineering community, as demonstrated in Fig.~\ref{fig:number_of_publications}. 
However, there lacks a standardized set of criteria for evaluating spike sorting algorithms, which poses a challenge for researchers when comparing algorithms and hardware combinations to suit specific applications. In this paper, our specific contributions are as follows:

\begin{enumerate}
    \item We formulate a set of criteria and standardized datset for evaluating spike sorting algorithms and hardware;
    \item We present a case study using our criteria to benchmark a \ac{RRAM} \ac{IMC} and compare performance to a traditional \ac{CMOS} hardware implementation.
\end{enumerate}

\section{Related Work}
Table~\ref{table:overview} shows an overview of several noteworthy publications from the past decade, and the criteria each used for evaluation. While all studies evaluate the classification accuracy of their algorithm, only some also investigate the algorithm's performance over a variety of other criteria. Due to the relatively recent re-emergence of the spike sorting field, there is currently no consensus on a standardized method to use. Additionally, unlike other well established signal processing techniques, such as power spectral analysis, accuracy is not the only criteria of consideration due to the highly demanding nature of spike sorting applications and experimental settings.

We note that this paper is not intended to serve as a survey or review paper like~\cite{gibson2008comparison}, and that while many of the formalized criteria have been previously reported in related works, as a collective, they have not been done so systematically.

\begin{table*}[!t]
\centering
\caption{Overview of criteria and algorithms used by spike sorting publications in the literature.}\label{table:overview}
\begin{tabu} to 1\textwidth {p{0.025\textwidth}p{0.025\textwidth}XX}
\toprule \toprule
\textbf{Paper} & \textbf{Year} & \textbf{Criterion(s)} & \textbf{Main Algorithm(s)} \\
\midrule
\cite{kim2000neural} & 2000 & Accuracy, noise tolerance. & Neural network spike classification. \\
\cite{quiroga2004unsupervised} & 2004 & Accuracy. & Wavelet feature extraction and superparamagnetic clustering. \\
\cite{zviagintsev2006algorithms} & 2006 & Accuracy, computational complexity, power consumption on custom hardware. & Integral Transform. \\
\cite{zhu2010fpga} & 2010 & Accuracy, custom \ac{FPGA} execution speed and precision, resources associated with each computing component. & Hardware implementation of neural network spike sorting on custom \ac{FPGA}. \\
\cite{balasubramanian2011fuzzy} & 2011 & Accuracy, alignment invariance. & Fuzzy logic spike sorter. \\
\cite{bestel2012novel} & 2012 & Accuracy. & Adaptable feature extraction. \\
\cite{kamboh2012computationally} & 2012 & Accuracy, computational complexity, power consumption on custom hardware. & \acp{ZCF} and hardware architecture. \\
\cite{karkare201375} & 2013 & Power consumption on custom system architecture, memory requirement, accuracy. & 16-channel online spike sorting algorithm and implantable hardware design. \\
\cite{caro2018spike} & 2018 & Accuracy, execution time. & Spike sorting based on shape, phase, and distribution features, and K-Tops clustering. \\
\cite{laboy2019normalized} & 2019 & Accuracy, firing rate and noise tolerance, requirement for manual intervention. & Normalized template matching for spike sorting. \\
\cite{shaeri2020framework} & 2020 & Accuracy, memory requirement. & Salient feature extraction and on-implant module design. \\
\cite{racz2020spike} & 2020 & Accuracy. & Spike sorting with \ac{DL}. \\
\cite{yang2014computationally} & 2020 & Noise tolerance, accuracy, computational complexity. & Feature extraction with feature denoising filter preserve maximum information. \\
\bottomrule \bottomrule
\end{tabu}
\end{table*}

\section{Preliminaries}
\begin{figure}[!t]
    \centering
    \includegraphics[width=1\linewidth]{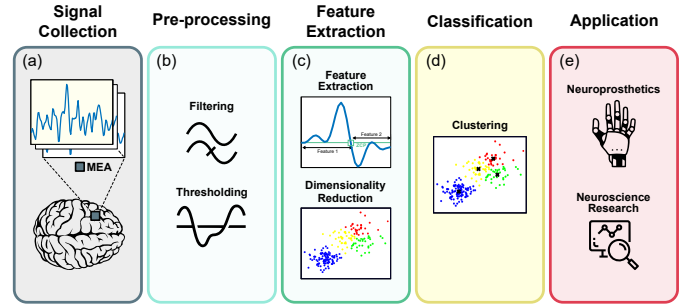}
    \caption{A general processing pipeline for spike sorting.}
    \label{fig:overview}
\end{figure}

Spike sorting refers to algorithms that detect individual spikes (action potentials) from extracellular neural recordings and classifies them according to their shapes, which attributes detected spikes to the originating neurons. This technique operates on the principal that different neurons tend to produce spikes of varying shapes, due to their varying proximity to the electrode, as well as their varying morphology of dendritic trees~\cite{gold2006origin}. Current spike sorting techniques generally involve 3 steps: 1) Spike detection; 2) Feature extraction; 3) Classification. A general processing pipeline is shown in Fig.~\ref{fig:overview}. Bandpass filtering is generally used to eliminate high frequency artefacts and low frequency noise. For online applications, a causal filter is required, while non-causal filters are preferred for offline analysis. Spike detection isolates single spike waveforms, typically of duration 1-3ms, and aligns them accordingly. Next, 2 or 3 features that best distinguish between different spike classes are extracted, which alleviates the problem of "curse of dimensionality". The extracted features are then used as inputs to supervised or unsupervised classification algorithms which output the corresponding class for each spike.

\section{Proposed Set of Formulated Criteria}
In Table~\ref{table:proposed_critera}, our proposed spike-sorting criteria is summarized. In this section, we discuss each criterion in more detail.

\begin{table}[!t]
\caption{Proposed spike-sorting criteria.}\label{table:proposed_critera}
\begin{tabu} to 0.5\textwidth {XX}
\toprule \toprule
\textbf{Criterion} & \textbf{Description} \\
\midrule
\circled{1} Detection Accuracy (\%) & Spike detection accuracy. \\
\circled{2} Detection AUROC & Spike detection Area Under the \ac{ROC} (AUROC). \\
\circled{3} Feature Extraction and Classification Accuracy (\%) & The overall sorting accuracy out of the correctly detected spikes.  \\
\circled{4} \ac{ICV} & Measurement of the compactness of each cluster. \\
\circled{5} Power (W/Ch) & Power (per channel). \\
\circled{6} Energy (J/Ch) & Energy (per channel). \\
\circled{7} Area (m$^2$/Ch) & Area (per channel). \\
\circled{8} Latency (s/Ch) & Latency (per channel).\\
\bottomrule \bottomrule
\end{tabu}
\end{table}

\subsection{Accuracy Performance}
Accuracy is the most commonly used criterion across almost all spike sorting publications, as it is the most direct indication of an algorithm's performance. As previously mentioned, most spike sorting algorithms are comprised of 3 stages: spike detection, feature extraction, and clustering. Amongst these, spike detection is often isolated and its accuracy (the \circled{1} spike detection accuracy) is evaluated independent of later steps. Likewise, feature extraction and clustering are often combined and tested for classification accuracy, independent of detection accuracy. This accuracy evaluation scheme is to ensure that the user can accurately determine how each stage contributes to the accuracy and make necessary changes if required. Additionally, noise tolerance metrics should be included to assess performance degradations.

\subsubsection{AUROC}
In addition to reporting the spike detection accuracy, many works also construct \ac{ROC} curves and report the \circled{2} detection AUROC for discrimination evaluation \cite{gibson2008comparison}.

\subsubsection{Feature Extraction and Classification Accuracy}
To eliminate any confounding effects from the spike detection step, when evaluating the \circled{3} feature extraction and classification accuracy, ground truth spikes should be used if they are available. 
When ground truth labels are not available, in lieu of the classification accuracy, the \circled{4} \ac{ICV}~\cite{regalia2016framework} metric, as defined in (\ref{eq:ICV}), can be used. 
\begin{equation}\label{eq:ICV}
\mathrm{ICV}=\frac{1}{N_{i}} \sum_{j=1}^{N_{i}}\left(v_{j}-\mu_{i}\right)^{2}
\end{equation}
The \ac{ICV} metric was originally used to measure the compactness of each cluster, where $v_{j}$ is the $j$th spike in the $i$th cluster, $\mu_{i}$ is the mean waveform, and $N_{i}$ is the number of spikes in cluster $i$.
For more comprehensive accuracy testing, performance on real data sets can be evaluated.

\subsubsection{Noise Tolerance}
Noise tolerance is crucial to consider when choosing spike sorting algorithms, as noise can significantly hinder some algorithms while others remain relatively more robust \cite{yang2014computationally}. Different datasets used by various studies have vastly different noise levels presenting challenges to direct comparison, hence a standardized dataset with varying noise level should be used such as the one used in this work. 

\begin{figure*}[!t]
    \centering
    \includegraphics[width=1\textwidth]{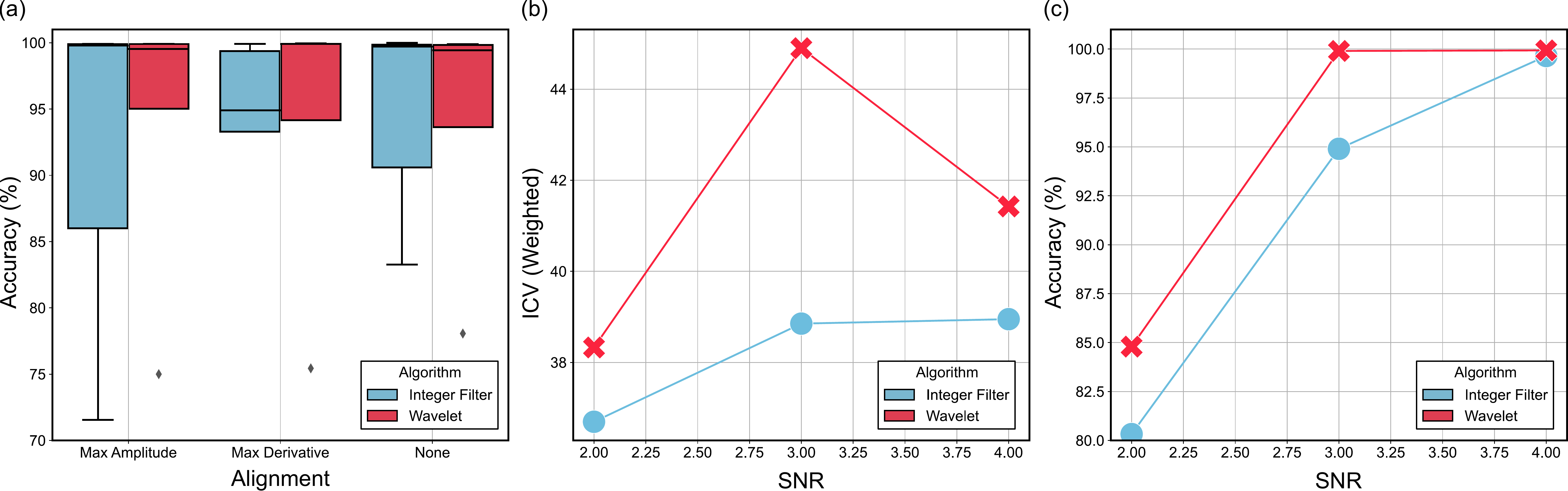}
    \caption{(a) The impact of spike alignment techniques on classification accuracy for both algorithms. It can be seen that alignment slightly improves the wavelet but hinders the integer filter technique. (b-c) A comparison of the \ac{ICV} and accuracy metrics when both algorithms are used for different \acp{SNR}. Between \ac{ICV} and accuracy metrics, a \ac{PCC} of 0.7232 is reported, which indicates weak correlation.}
    \label{fig:results}
\end{figure*}

\subsubsection{Alignment Requirements}
Alignment is an additional spike sorting step that refers to aligning spikes before the feature extraction and classification. Some algorithms benefit significantly from this additional step, while some do not \cite{balasubramanian2011fuzzy}. Alignment benefits should be compared when evaluating \circled{1}, \circled{2}, \circled{3}, and \circled{4}.

\subsection{Power, Energy, Area, and Latency}
The \circled{5} power, \circled{6} energy, \circled{7} area, and \circled{8} latency (per channel) should be reported for all stages (a)-(e) in Fig.~\ref{fig:overview} for a specific hardware implementation technology. In addition to facilitating direct comparison, these metrics can be used to evaluate the online applicability of a given system under different resource constraints.

\begin{table}[!t]
\caption{Hardware performance of our simulated system adopting the \ac{DWT} for feature extraction. $^\diamond$Not reported.}\label{table:reported_metrics}
\begin{tabu} to 0.5\textwidth {XX[r]X[r]}
\toprule \toprule
\textbf{Criterion} & \textbf{Our Reported Values} & \textbf{65-nm CMOS~\cite{do2018area}}\\
\midrule
Power (mW/Ch) & 10.72 & 0.000175\\
Energy (mJ/Ch) & 1.45 & N/R$^\diamond$\\
Area (m$^2$/Ch) & 0.66 & 4.14e-7 \\
Latency (ms/Ch) & 135.53 & N/R$^\diamond$\\
\bottomrule \bottomrule
\end{tabu}
\end{table}

\section{The \ac{SSOER} Dataset}\label{sec:SOER_dataset}
One of the greatest challenges facing spike sorting algorithm development is the lack of labelled experimental data, which gives researchers the ability to validate their algorithms and measure evaluate performance. Hence, synthetic extracellular recordings have been developed to simulate neural recordings, constructed from known spike shapes as ground truths. In this paper, we formulate a dataset for unsupervised and supervised spike sorting algorithms entitled \ac{SSOER}~\cite{Zhang2022}, which is the amalgamation of five smaller synthetic datasets with varying \acp{SNR}, which were originally presented in~\cite{martinez2009realistic}. These datasets are comprised of spikes from a database with 594 different average spike shapes, taken from real recordings from monkey neocortex and basal ganglia. Recordings are sampled with a sampling frequency of 96 kHz, filtered, and then down-sampled to 24 kHz. \ac{SSOER} is made openly accessible\footnote{\url{https://github.com/TimothyZh/MWSCAS2022SpikeSortingCriteria}}.

\section{Case Study: A \ac{RRAM} \ac{IMC} System}\label{sec:results}
To demonstrate the robustness of our formalized approach, we present a case study evaluating a simulated \ac{RRAM} \ac{IMC} system implemented using a 22nm FDSOI \ac{CMOS} process with device integration at the \ac{BEOL}. \ac{RRAM} devices can be arranged in crossbar configurations to efficiently perform analogue \acp{VMM} in-memory~\cite{amirsoleimani2020memory,Lammie2020,RahimiAzghadi2020}, which is the most dominant operation in many popular algorithms. For spike-sorting applications, \ac{RRAM} is preferable over charge-based memory such as SRAM and DRAM due to its scalability down to nanometer scale \cite{sebastian2020memory}. We use a crossbar model proposed by Primeau et al. 2021~\cite{Amirsoleimani2021}, which is based on existing semi-passive crossbar models~\cite{chen2013comprehensive,Lammie2020}, to simulate the feature extraction stage of the spike sorting pipeline. To compute the \ac{DWT}, five unique decomposition levels (iterations) were used, each comprising of many \ac{VMM} operations. These were mapped onto a singular crossbar made-up of 64 modular tiles of (8 $\times$ 8) \ac{RRAM} devices. More comprehensive system- and circuit-level information, all simulated models, and detailed hardware evaluation
methodologies are made accessible$^1$.
In Fig.~\ref{fig:results}, the performance of our \ac{IMC} implemented \ac{DWT}, i.e., metrics \circled{1} -- \circled{4}, is compared against a digital integer filter algorithm from \cite{do2018area}. Metrics \circled{5} -- \circled{8} are reported in Table~\ref{table:reported_metrics}.

From Fig.~\ref{fig:results} (a), for both algorithms, it can be observed that alignment improved classification accuracy across all datasets in most cases, while integer filter generally benefited more from the alignment. In Figs.~\ref{fig:results} (b-c), a \ac{PCC} of 0.7232 was reported between both metrics, indicating weak correlation. This, however, is still deemed significant, considering that the \ac{ICV} metric can be determined without the need to label data. The classification accuracy decreased for both algorithms with increasing noise level, less so with the wavelet technique.

In Table~\ref{table:reported_metrics}, the total reported power and area values of the simulated \ac{IMC} \ac{RRAM} system are significantly larger than that reported by~\cite{do2018area}. As this case study was designed to demonstrate the effectiveness of our formalized approach, and not to investigate the feasibility of \ac{IMC} \ac{RRAM} systems for spike sorting applications, no architectural hardware optimizations were performed, such as gating and latency balancing.

\section{Conclusion}
In this work, a formalized approach for spike sorting algorithms and hardware evaluations was proposed and a case study has been performed to demonstrate the efficacy of such methodology. With the consolidated \ac{SSOER} dataset, the critical challenge of direct comparison between systems have been addressed which should aid future researchers with selecting the appropriate spike sorting systems as well as identifying areas for improvements. 

\bibliographystyle{IEEEtran}
\bibliography{References}

\end{document}